\documentclass{article}

\usepackage{PRIMEarxiv}

\usepackage[utf8]{inputenc} 
\usepackage[T1]{fontenc}    
\usepackage{wrapfig}
\usepackage{booktabs}       
\usepackage{amsmath}
\usepackage{natbib}
\usepackage{shortvrb}

\usepackage{bm}
\usepackage{subcaption}
\usepackage{tikz}
\usetikzlibrary {arrows.meta} 
\usepackage{comment}

\usepackage{tikz}

\DeclareMathOperator*{\argmax}{arg\max}

\newcommand{\Real}{\mathbb{R}}

\newcommand{\indep}{\perp \!\!\! \perp}

\newcommand{\CI}{\operatorname{CI}}

\newcommand{\MSE}{\operatorname{MSE}}

\usepackage[utf8]{inputenc} 
\usepackage[T1]{fontenc}    

\usepackage{graphicx}
\usepackage{amsmath}
\usepackage{amssymb}
\usepackage{booktabs}

\usepackage{subcaption}
\usepackage{fourier}

\graphicspath{{figures/}}     

\usepackage[colorlinks,citecolor=blue]{hyperref}
\usepackage{times}

\usepackage[utf8]{inputenc} 
\usepackage[T1]{fontenc}    
\usepackage{wrapfig}
\usepackage{booktabs}       
\usepackage{amsmath}
\usepackage{bm}
\usepackage{subcaption}
\usepackage{tikz}
 \usetikzlibrary {arrows.meta} 
\usepackage{comment}

\title{Recovering Latent Confounders from High-dimensional Proxy Variables}

\author{%
Nathan Mankovich$^\ast$$^\#$, Homer Durand$^\#$, Emiliano Diaz$^\#$, Gherardo Varando$^\#$,  and Gustau Camps-Valls \\
Image Processing Laboratory (IPL)\\
Parc Científic Universitat de València\\
46980 Paterna (València). Spain\\
$^\ast$\texttt{nathan.mankovich@gmail.com} \\
$^\#$ denotes equal contribution
}

\begin{document}

\maketitle

\begin{abstract}
    Detecting latent confounders from proxy variables is an essential problem in causal effect estimation. Previous approaches are limited to low-dimensional proxies, sorted proxies, and binary treatments. We remove these assumptions and present a novel Proxy Confounder Factorization (PCF) framework for continuous treatment effect estimation when latent confounders manifest through high-dimensional, mixed proxy variables. 
    For specific sample sizes, our two-step PCF implementation, 
    using Independent Component Analysis (ICA-PCF), and the end-to-end implementation, using Gradient Descent (GD-PCF), achieve high correlation with the latent confounder and low absolute error in causal effect estimation with synthetic datasets in the high sample size regime. Even when faced with climate data, ICA-PCF recovers four components that explain $75.9\%$ of the variance in the North Atlantic Oscillation, a known confounder of precipitation patterns in Europe. Code for our PCF implementations and experiments can be found \href{https://github.com/IPL-UV/confound_it}{here}
    The proposed methodology constitutes a stepping stone towards discovering latent confounders and can be applied to many problems in disciplines dealing with high-dimensional observed proxies, e.g., spatiotemporal fields.
\end{abstract}
\vspace{0.25cm}

\section{Introduction}

Estimating the causal effect between a \emph{cause} (\emph{treatment}) variable $X$ and the \emph{output} (\emph{outcome}) variable $Y$ is a central problem in causal inference~\citep{Peters2017, hernan2023causal} with vast applications in many fields of science~\citep{glass2013causal,Nature-2023RungeGVE}. When using observational data, adjusting for confounding variables (\emph{confounders}) is a crucial component of causal inference. Unaccounted confounders could introduce bias and void inference results. Often, we observe data from high-dimensional proxy variables (\emph{proxies}) for low-dimensional and latent (unobserved) confounders. Proper adjustment for these confounders is necessary for correct causal effect estimation. In this paper, we address the Proxy Confounder Factorization (PCF) problem of latent confounder detection from high dimensional proxies and use these confounders to perform adjusted causal inference between treatment and outcome with the novel PCF framework. Previous approaches for the PCF problem often simplify the problem to low dimensional proxies, solely causal effect estimation without detecting confounders, and/or only binary treatment variables. Moreover, many of these methods rely on intricate (neural) network models. 

First, the simplification of measuring low dimensional proxies for detecting confounding variables and performing adjusted causal effect estimation has been addressed as a form of measurement bias in causal inference~\citep{Pearl10, Biometrika-2014KurokiP}. This simplification is known as the field of proximal causal learning, first introduced by~\cite{Biometrika-2018MiaoGT} and~\cite{Arxiv-2020TchetgenYCSM}. Proximal causal learning aims to perform causal inference in observational studies without the stringent and often criticized assumption of {\em exchangeability}~\citep{hernan2023causal}. Similar to our setting, these methods consider observed data only from the treatment, effect, and proxies for the confounders. Unlike our setting, the standard literature in proximal causal learning assumes to have access to three separate groups of proxies: 
(a) common causes of $X$ and $Y$; 
(b) those only related to $X$ (treatment-inducing proxies); and 
(c) those only related to $Y$ (outcome-inducing proxies). 
A body of literature addresses the theoretical side of this problem along with some practical examples~\citep{Biometrika-2018MiaoGT, Arxiv-2020TchetgenYCSM, ICML-2021MastouriZGK, OpenReview-2023SverdrupC, JASA-2023CuiPSM}.

Outside of proximal causal learning, most works restrict their setting to binary treatments and generally rely heavily on complex network-based methods. Two of these methods that focus on binary treatments and are closest to our work are~\cite{wu2022learning,chu2021learning}. However,~\cite{wu2022learning} does not estimate latent confounders and requires instrumental and adjustment variables to have the same dimensions. Several additional studies employ intricate loss functions for supervised models to estimate treatment effects while occasionally identifying latent confounding variables~\citep{louizos2017causal, hassanpour2019learning, chu2021learning, wu2022learning, PDS-2022-WyssYCB, cheng2022learning}. Moreover, many of these approaches simplify the problem by considering only binary treatment variables~\citep{louizos2017causal, luo2020matching, chu2021learning, cheng2022learning, wu2022learning, cheng2022sufficient}.
 
The harder PCF problem appears in many scientific disciplines, e.g., neuroscience, Earth, or vision sciences \cite{Nature-2023RungeGVE,snoek2019control,deshpande2022deep}. In the past, similar problems that primarily focus on the simplified case of estimating treatment effects while indirectly detecting confounding variables have been prevalent in the areas of biology and epidemiology~\citep{AJE-2015-FranklinEGS, shortreed2017outcome, chu2021learning, PDS-2022-WyssYCB}. Outside of biology, Earth and climate sciences deal with complex data representing spatio-temporal interactions among variables. For example, distant and lagged correlations between weather patterns in different regions are called teleconnections~\cite{Wallace1981TeleconnectionsIT}. A variant of principal component analysis (PCA, a.k.a. empirical orthogonal function EOF) called Varimax PCA is used to extract causally meaningful variables for investigating atmospheric teleconnections and general dynamics~\citep{Runge2015}. While correlation and information/ entropy transfer have been used to study these links, causality provides a more precise way to understand them. Our PCF problem of latent confounder detection from high dimensional proxies and causal effect estimation fits nicely into the flowchart outlined in~\cite{Nature-2023RungeGVE}.

To address the PCF problem of low-dim\-ens\-ional confounder detection from high dimensional proxies and adjusted causal effect estimation between continuous treatment and outcome, we introduce the PCF framework. PCF essentially performs a dimensionality reduction (DR) of the measured proxy to extract the \emph{most confounding} latent variable(s).  We implement and analyze PCF using (i) \emph{reduction and selection:} off-the-shelf DR algorithms paired with basic linear regression $t$-tests and (ii) \emph{end-to-end:} a method trained to optimize the regression errors and conditional independence constraints (similar to~\cite{chu2021learning}). The off-the-shelf DR algorithms considered in (i) are PCA~\citep{JEP-1933Hotelling}, Partial Least Squares (PLS)~\citep{Elsevier-1986GeladiK}, and Independent Component Analysis (ICA)~\citep{jutten1991blind, hyvarinen2000independent} and we call the associated frameworks PCA-PCF, PLS-PCF, and ICA-PCF. The end-to-end optimized method, (ii), is called GD-PCF and is tailored to the assumptions of our PCF framework. We illustrate the performance of PCF in terms of accuracy and efficacy on synthetic data and a challenging climate science problem. Specifically, we employ Atlantic pressure data and precipitation records from Denmark (DEN) and the Mediterranean (MED) to pinpoint a confounder, the North Atlantic Oscillation (NAO). Subsequently, we conduct adjusted causal effect estimation for precipitation between both regions~\citep{BAMS-2021-KretschmerAAP}. \emph{Overall, the PCF problem removes the binary treatment assumption and the lack of interpretability of neural networks, considers the case of high-dimensional confounding proxies, and the PCF framework detects known climate phenomena.} The main contributions of this work are 
\begin{itemize}
    \item Formulate the PCF problem which addresses the recovery of latent confounders from high dimensional proxies considering a continuous treatment variable.
    \item Evaluate several DR methods within the PCF framework and discuss their implications.
    \item Illustrate the performance in both synthetic and challenging, real problems. 
\end{itemize}


\section{Methods}\label{sec:methods}

We formalize the PCF problem of recovering low-dimensional latent confounders from high-dimensional proxies and performing adjusted causal effect estimation. Then, we provide the assumed structural causal model (SCM), explain our implementations of the PCF framework, and formalize the evaluation metrics for our methods. 

\paragraph{Notation} 
$X$ denotes a random variable (RV). $x \in \mathbb{R}$ denotes a sample and  $\mathbf{x} \in \mathbb{R}^n$ refers to s vector of $n$ samples of $X$. The matrix of $n$ samples of a $p$-dimensional RV $X = (X_1,X_2,\dots, X_p)$ is denoted $\mathbf{X} \in \mathbb{R}^{n \times p}$. We denote the $j$th column of $\mathbf{X}$ as $\mathbf{x}_j$ or $(\mathbf{X})_j$ and the $i,j$th entry of $\mathbf{X}$ as $\mathbf{X}_{i,j} = x_j^{(i)}$. Finally, $\mathbb{P}$ is a probability distribution. Some specific symbols corresponding to the PCF problem are named in Table~\ref{tab: symbols}.

\begin{table}[ht!]
    \centering
    \caption{Definitions of key notation for PCF.}
    \begin{tabular}{cc}
        \toprule
        Symbol & Definition \\
        \midrule
        $X$ & Treatment\\
        $Y$ & Outcome\\
        $\alpha \in \mathbb{R}$ & Causal coefficient of $X$ on $Y$\\
        $U = (U_1, U_2, \dots, U_p)$ & High dimensional proxy\\
        $Z = (Z_x, Z_c, Z_y, Z_n)$ & Potential confounders\\
        $Z_c$ & Latent confounders\\ 
        \bottomrule
    \end{tabular}
    \label{tab: symbols}
\end{table}

\subsection{Problem statement}\label{sec:scm}
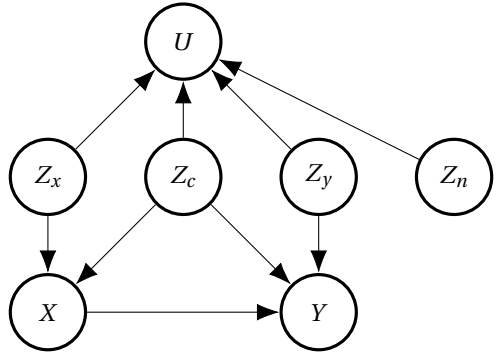
\begin{figure}[ht!]
    \centering
    \begin{tikzpicture}[auto, scale=12,
    node distance = 18mm and 6mm,
    edge/.style ={arrows=-{Latex[length=3mm]}},
    dashededge/.style ={dashed}, 
	node/.style={circle,inner sep=1mm,minimum size=1cm,draw,black,very thick,text=black}]

     \node [node] (x)  {$X$};
     \node [node] (y) at (0.3,0) {$Y$};
     \node [node] (zx) [above of = x] {$Z_x$};
     \node [node] (zy) [above of = y]  {$Z_y$};
     \node [node] (zc) [right of=zx] {$Z_c$};
     \node [node] (zn) [right of=zy] {$Z_n$};

     \node [node] (proxy) [above of=zc]  {$U$};

     \path[edge] (zc) edge (x);
     \path[edge] (zc) edge (y);
     \path[edge] (zc) edge (proxy);
     
     \path[edge] (x) edge (y);
     \path[edge] (zx) edge (x);
     \path[edge] (zy) edge (y);
     \path[edge] (zx) edge (proxy);
     \path[edge] (zy) edge (proxy);
     \path[edge] (zn) edge (proxy);
    \end{tikzpicture}
    \caption{DAG studied in this work and induced by the Structural Causal Model (SCM) in Eq.~\eqref{eq:SCM}.}
    \label{fig:dag_scm}
\end{figure}

We consider the PCF problem of estimating the true causal coefficient ($\alpha$) of the treatment $X$ on the outcome $Y$ while controlling for a low-dimensional latent confounder $Z_c$ that is estimated using observations from a high-dimensional confounder-proxy $U$\footnote{In general, estimating $\alpha$ from observational data of $X$ and $Y$ alone is impossible (even with infinite data) since unobserved confounding could bias the estimation of the coefficient.}. We additionally assume that $U$ is a proxy for a set of potential confounding variables: $Z = (Z_x, Z_c, Z_y, Z_n)$, which separates into the instrumental $Z_x$, confounder $Z_c$, adjustment $Z_y$, and `useless' $Z_n$ variables. The assumed causal relationships from this problem are summarized in the directed acyclic graph (DAG) in Fig~\ref{fig:dag_scm}. 
We use this DAG and assume linear relationships between random variables. Using these assumptions, we define the distribution of $(X, Y, Z, U)$ by following linear Structural Causal Model (SCM):
\begin{align}\label{eq:SCM}
\begin{aligned}
    Z &:= (Z_x, Z_c, Z_y, Z_n) \sim \mathbb{P}_{Z}\\
    U &= Z\mathbf{W} + N_u\\
    X &= Z_x \mathbf{a}_x + Z_c \mathbf{a}_c + N_x\\
    Y &= \alpha X + Z_c \mathbf{b}_c + Z_y \mathbf{a}_y + N_y\\
    (N&_u, N_x, N_y) \sim \mathbb{P}_{N}\\
\end{aligned}
\end{align}
We make three assumptions about our SCM: 
(1) $Z$ is of dimension $k \ll p$, 
(2) $Z, N_x, N_y, N_u$ are mutually independent, and 
(3) $Z, N_x, N_y, N_u$ are not observed. 
In summary, we use observations of $U$, $X$, and $Y$ to perform the following inter-related tasks outlined by the PCF problem: (i) estimate $Z_c$ and (ii) estimate the causal effect of $X$ on $Y$ adjusting for $Z_c$. 

\subsection{PCF Implementation}  
We offer two types of PCF implementations: one set with reduction and selection steps and an end-to-end method optimized by gradient descent. 

\paragraph{PCF with reduction and selection} 
PCF, with reduction and selection, first reduces samples of $U$ to estimated samples of $Z$, then selects $Z_c$. Our implementations of PCF with reduction and selection utilize PCA, PLS, and ICA for the \emph{reduction} step. Using a particular DR method constrains the form of the map from $\mathbf{Z}$ to the proxies $\mathbf{U}$. PCA seeks to maximize the variance of the projections, 
PLS maximizes the covariance between the features and the target variable, 
and ICA obtains independent signals \citep{Arenas13}. Specifically, PCA and PLS find approximate right inverses of the mapping from our SCM (Eq.~\eqref{eq:SCM}) and restrict this mapping to be a truncated orthogonal matrix. ICA directly estimates the mapping and $\mathbf{Z}$ by obtaining independent signals that are identifiable up to a scaling and random permutation only when no more than one column of $\mathbf{Z}$ comes from a Gaussian distribution and $n \geq k$~\citep{SP-1994Comon, Springer-1998LeeL}. In the small sample size ($n$) regime, ICA is known to empirically perform similarly to whitened PCA \citep{PR-2012DengLH}. More information on the particularities of PCA, PLS, and ICA is in Appendix~\ref{app:dr}. 

For the \emph{selection} step, we use regression to test each DR-detected potential confounder individually. For each candidate confounder (the estimated columns of $Z$, denoted $\widehat{\mathbf{z}}_i$), we fit two linear regression models predicting $X$ from $\widehat{\mathbf{z}}_i$ and another predicting $Y$ from $\widehat{\mathbf{z}}_i$ and $X$, according to the DAG in Fig.~\ref{fig:dag_scm}. We then compute the two $p$-values $p_x$ and $p_y$ corresponding to the two null hypotheses that the coefficients of $\widehat{\mathbf{z}}_i$ are $0$ in the two regression models, respectively. 
Finally, we select the component $ \widehat{\mathbf{z}}_i$, which minimizes the sum of $ p$-values ($ p_x + p_y$) as the estimator of samples of the true latent confounder, $\mathbf{z}_c$. According to their respective DR algorithms, we refer to PCF with these reduction and selection methods as PCA-PCF, PLS-PCF, and ICA-PCF.

\paragraph{The end-to-end PCF algorithm (GD-PCF)}
Gradient descent PCF (GD-PCF) is a methodology optimized for the PCF problem. It approaches the problem from a machine learning perspective and leverages the DAG in Fig.~\ref{fig:dag_scm} by imposing conditional independence among variables through the Hilbert-Schmidt independence criterion (HSIC). Although this approach aligns with several other works~\cite{Greenfeld2019,Diaz_2023}, this method stands out as a distinctly novel algorithm tailored to address the PCF problem. 

GD-PCF is implemented by (1) parameterizing the sought-after latent variables $Z$ as linear functions of the proxy $U$ and (2) parameterizing $X$ and $Y$ as linear functions of $Z$ as prescribed by the assumed DAG in Fig.~\ref{fig:dag_scm}. Parameters related to the models for $X$ and $Y$ are set to the closed ridge regression form but depend implicitly on the models for $Z$. Parameters related to the models for $Z$ are optimized simultaneously with stochastic gradient descent (GD) and automatic differentiation with respect to a loss composed of a risk term and a regularizer. The risk term is the mean squared error for $X$ and $Y$ regressions, while the regularizer penalizes violations of the conditional independence relations implied by DAG in Fig.~\ref{fig:dag_scm}.

To simplify the explanation of GD-PCF, we assume $X$, $Y$, $Z_x$, $Z_y$, and $Z_c$ are all $1$-dimensional. The following model is easy to generalize to higher dimensional variables. The GD-PCF algorithm uses five linear models to estimate $\mathbf{x},\mathbf{y},\mathbf{z}_x, \mathbf{z}_y$ and $\mathbf{z}_c$. 
\begin{align*}
\begin{aligned}
 \mathbf{z}_x &= \mathbf{U} \mathbf{v}_x \\
 \mathbf{z}_y &= \mathbf{U} \mathbf{v}_y \\
 \mathbf{z}_c &= \mathbf{U} \mathbf{v}_c \\
 \mathbf{x} &= \mathbf{z}_xa_x + \mathbf{z}_ca_c \\
 \mathbf{y} &= \mathbf{x}\alpha+\mathbf{z}_yb_y + \mathbf{z}_cb_c,
 \end{aligned}
\end{align*}
where $\mathbf{v}_x, \mathbf{v}_y, \mathbf{v}_c \in \Real^p$ and $a_x, a_c, \alpha, b_y, b_c \in \Real$. We follow a similar modeling strategy to~\citep{Diaz_2023} where models for observed variables are estimated using (kernel) ridge regression such that the loss only depends on the latent variable model parameters. Parameters $a_x, a_c, \alpha, b_y, b_c$ are estimated using the closed ridge regression form for regressing $\mathbf{b}$ on $\mathbf{A}$:
\begin{align*}
\bm{\gamma} = (\mathbf{A}^T \mathbf{A}+ \lambda \mathbf{I})^{-1}\mathbf{A}^T \mathbf{b},
\end{align*}
where for the $\mathbf{x}$-model $\mathbf{A}:=[\mathbf{z}_x, \: \mathbf{z}_c]^T$, $\mathbf{b}:=\mathbf{x}$, and $\bm{\gamma}:=[a_x, \: a_c]^T$ whereas for the $\mathbf{y}$-model $\mathbf{A}:=[\mathbf{x}, \: \mathbf{z}_y, \: \mathbf{z}_c]^T$, $\mathbf{b}:=\mathbf{y}$, and $\bm{\gamma}:=[\alpha, \: b_y, \: b_c]^T$. 
With this choice, we only need to estimate $\mathbf{v}_x, \mathbf{v}_y$ and $\mathbf{v}_c$. These are estimated by optimizing the reconstruction error of $X$ and $Z$ such that violations of the conditional independence relations of the assumed DAG are penalized. The relevant conditional independence relations implicit in the DAG in Fig.~\ref{fig:dag_scm} are the following: $Z_x \indep Z_y$, $Z_x \indep Z_c$, $Z_y \indep Z_c$, $Z_y \indep X$ and $Z_x \indep Y|X$.

The loss function is composed of two terms:
\begin{equation}
\mathcal{L}(\mathbf{V}';\mathbf{x},\mathbf{y},\mathbf{U}) = \gamma\log \MSE(\mathbf{V}';\hat{\mathbf{x}}, \hat{\mathbf{y}}) + \eta \log \CI(\mathbf{V}',\hat{\mathbf{x}},\hat{\mathbf{y}},\hat{\mathbf{Z}}),
\end{equation}
where $\mathbf{Z} = [\mathbf{z}_x, \mathbf{z}_c, \mathbf{z}_y]$
and $\mathbf{V}':=[\mathbf{v}_x, \mathbf{v}_y, \mathbf{v}_c]^T$. 
The mean squared error (MSE) term is decomposed as follows: 
\begin{equation}
\MSE(\mathbf{V}';\hat{\mathbf{x}}, \hat{\mathbf{y}}) = \frac{1}{n}\|\mathbf{y}-\hat{\mathbf{y}} \|_2^2+\frac{1}{n}\|\mathbf{x}-\hat{\mathbf{x}}\|_2^2.
\end{equation}
We use the normalized HSIC~\citep{HSIC} between variables $X$ and $Y$ (with empirical estimate denoted $\text{nHC}(\mathbf{x},\mathbf{y})$) to measure the degree to which the statement $X\indep Y$ is true. Since we are in an additive setting to test if $Z\indep Y|X$, we may regress $Y$ on $X$ and check if there is any dependence left between the regression residuals $\mathbf{r}_{X\rightarrow Y}:=\mathbf{y}-\hat{\mathbf{y}}(\mathbf{x})$ and $Z$: $\text{nHC}(\mathbf{z},\mathbf{r}_{y\rightarrow x})$. As such, the conditional independence term is decomposed as follows:
\begin{equation*}
\CI(\mathbf{V}';\hat{\mathbf{x}},\hat{\mathbf{y}},\hat{\mathbf{Z}}) = \text{nHC}(\mathbf{z}_x, \mathbf{z}_y)+\text{nHC}(\mathbf{z}_x, \mathbf{z}_c)+\text{nHC}(\mathbf{z}_y, \mathbf{z}_c)\\ +\text{nHC}(\mathbf{z}_y, \mathbf{x})+\text{nHC}(\mathbf{z}_x, \mathbf{r}_{X \rightarrow Y}).
\end{equation*}
We optimize using stochastic gradient descent with a batch size of $\lfloor \max(0.1n, 25) \rfloor$ and a learning rate of $0.001$. We performed $3000$ gradient updates. We used the following hyperparameters: $\lambda=0.01$, $\gamma=1$, $\eta=1$. The algorithm was implemented using the Python package JAX \citep{jax2018github}. 

\subsection{Metrics}
To evaluate the methods, we employ a diverse set of metrics. First, to gauge our ability to recover the latent confounder up to sign and scaling, we utilize the Absolute Correlation ({AbsCor}) metric, comparing our estimation to the real confounder $Z_c$. For evaluating our causal effect estimate ($\hat{\alpha}$), we employ both the Absolute Error ({AE}) and the Absolute Error Ratio ({AER}): AE quantifies the difference between $\hat{\alpha}$ and the causal effect in the SCM ($\alpha$) using $\text{AE} = |\alpha - \hat{\alpha}|$, whereas AER is normalized for the estimated causal effect using a baseline latent confounder estimate $\hat{\alpha}_0$, and thus $\text{AER}<1$ implies $\hat{\alpha}$ is a better estimate of $\alpha$ than $\hat{\alpha}_0$.


\section{Experiments}\label{sec:experiments}

We now investigate the efficacy of PCF by performing synthetic experiments with varying distributions, sample sizes, and proxy sizes and using it on climate science data.

\subsection{Synthetic experiments}
We generate synthetic data by sampling a $k=20$-dimensional $Z = (Z_x, Z_c, Z_y, Z_n)$ from four distributions: uniform, gamma, exponential, and Gaussian with $Z_x$ and $Z_y$ having $6$ dimensions and $Z_c$ having $1$ dimension. Then we compute samples for $X$, $Y$, and $U$ using the processes described in Sec.~\ref{sec:scm}, and we fix $p=1000$ and vary sample size ($n$). 

Specifically, we sample $\textbf{Z}$, $\textbf{U}$, $\textbf{x}$, and $\textbf{y}$ according to Eq.~\eqref{eq:SCM}. $\mathbf{Z}$ is sampled from an arbitrary distribution with a standard deviation of $1$. $\mathbf{W}$ is drawn from a standard normal distribution $\mathcal{N}(0,1)$. The SCM parameters $\mathbf{A_x}$, $a_c$, $\alpha$, $\mathbf{B_y}$, and $b_c$, are sampled from the uniform distribution $\mathcal{U}(0.5,1.5)$. The SCM noises, $\mathbf{N}_u$, $\mathbf{n}_x$, and $\mathbf{n}_y$, are sampled from $\mathcal{N}(0,1)$.

\begin{figure}[ht!]
    \centering
    \includegraphics{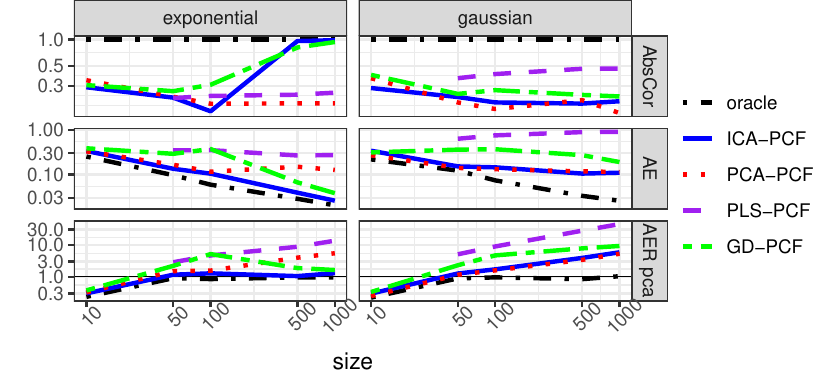}
    \caption{PCF implementations on synthetic data with exponential and Gaussian distributions for sampling $Z$. Performance is evaluated across different sample sizes in terms of correlation with the true confounder and estimating the causal effect. Metrics for the true confounder ($Z_c$) are labeled ``oracle.'' $x$-axis is log-scale and $y$-axes for AE and AER are log scale.}
    \label{fig:synthetic}
\end{figure}

\begin{figure}[ht!]
    \centering
    \includegraphics{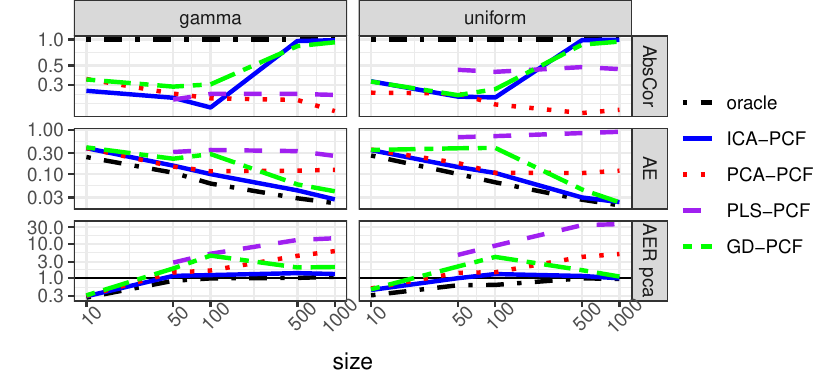}
    \caption{PCF implementations on synthetic data with gamma and uniform distributions for sampling $Z$. Performance is evaluated across different sample sizes in terms of correlation with the true confounder and estimating the causal effect. Metrics for the true confounder ($Z_c$) are labeled ``oracle.'' $x$-axis is log-scale and $y$-axes for AE and AER are log scale.}
    \label{fig:synthetic1}
\end{figure}

\begin{figure}[ht!]
    \centering
    \includegraphics{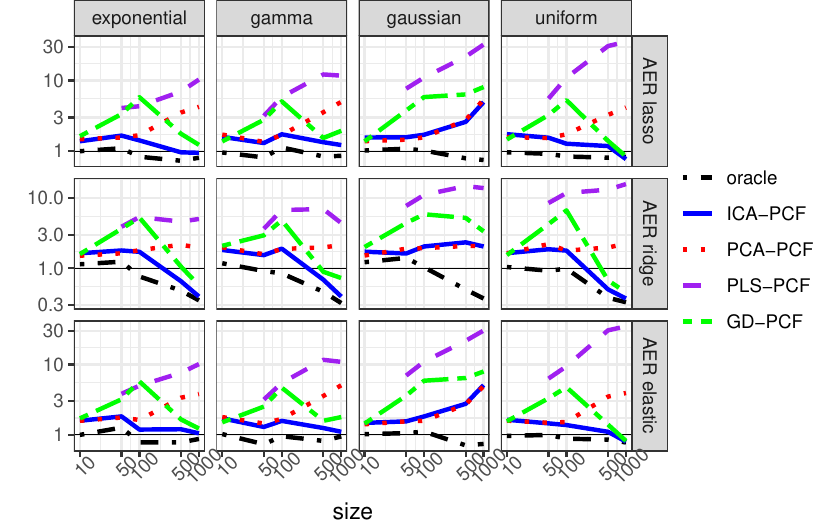}
    \caption{Relative performance of PCF implementations on synthetic data with respect to baselines (lasso, ridge, and elastic net) to estimate the causal coefficient. Values are plotted across different sample sizes. Metrics for the true confounder ($Z_c$) are labeled ``oracle.'' The $x$- and $y$-axes are in log-scale.}
    \label{fig:synthetic2}
\end{figure}

We perform sets of comparisons. The first compares our PCF implementations to the oracle estimate (see Fig.~\ref{fig:synthetic},~\ref{fig:synthetic1}), and the second compares to baseline causal coefficient estimation performed using regression variants controlling for the proxies (see Fig.~\ref{fig:synthetic2}).

\paragraph{Comparison between PCF methods} The median values of our metrics for $100$ trials are plotted in Fig.~\ref{fig:synthetic} for exponential and Gaussian distributions. 
Results for the other distributions are given in Fig.~\ref{fig:synthetic1} and show a similar pattern to the exponential case. In this case, AER is computed with $\hat{\alpha}_0$ computed using the first $k=20$ principal components of the proxies.

ICA-PCF is the best method for non-Gaussian distributions, especially for high sample sizes. Notice that the correlation approaches $1$, and the AE statistics almost match those of $Z_c$. This is because the ICA algorithm mirrors the assumed SCM (Eq.\eqref{eq:SCM}) by decomposing $U = Z \mathbf{W}$ and is known to accurately detect the signal ($Z$) and mixing matrix ($\mathbf{W}$) with non-Gaussian distributions and large enough sample sizes. So, any error from ICA-PCF in these cases should only be coming from the ``selection step'' using regression. Finally, the poor performance of ICA for Gaussian distributions is expected since ICA is known not to solve $U = Z \mathbf{W}$ for Gaussian distributions.

GD-PCF is initialized with ICA and is optimized using the properties of our assumed SCM (Eq.~\eqref{eq:SCM}). So we expect it to improve the ICA results. We find that GD-PCF exhibits higher correlations and AE statistics than ICA, suggesting that it has better $Z_c$ estimates but worse causal effect estimation. Perhaps this shortcoming can be addressed with further parameter tuning within GD-PCF.

On the other hand, we expect PCA and PLS not to recover $Z$ because they (i) assume the inevitability of $\mathbf{W}$ and (ii) estimate $Z$ by separating its dimensions according to maximum directions of variance and covariance of $U$ and $(X,Y)$ respectively. We see this reflected in the results because PLS did not converge for a sample size of $10$ and generally has a better correlation and worse AE statistics than PCA. PLS and PCA metrics remain constant across sample sizes, whereas the AER decreases as sample size increases. This decrease in performance is because the $\hat{\alpha}_{0}$ estimate of $\alpha$ improves as sample size increases. In the Gaussian case, PLS has the best correlation, and PCA has the best AE statistics.

\paragraph{Comparison to baselines} We compare the performance of PCF implementations to baseline estimates of the causal coefficient in Fig.~\ref{fig:synthetic2}. In this case, AER is computed with $\hat{\alpha}_0$ computed using the baseline regression (e.g., LASSO, ridge, or elastic net regression) for causal effect estimation controlling for all of the proxies. These baselines control for the latent confounders by regularizing the coefficients on the proxy contribution.

Overall, elastic net regression outperforms our PCF methods for causal coefficient estimation at all sizes and distributions other than $1000$ samples with $Z$ sampled from the uniform distribution. Similarly, LASSO regression outperforms all PCF implementations for all sample sizes for gamma and Gaussian distributions. But, for large sample sizes, and uniform or exponential distributions of $Z$, ICA-PCF (and GD-PCF only for uniform) produces a better causal coefficient estimation than the baselines. Also with enough samples, the estimated causal coefficients with ICA-PCF and GD-PCF are better than those estimated with ridge regression for all $Z$ distributions other than Gaussian. 

Overall, we notice that the baseline regressions outperform the oracle for casual coefficient estimation in the low sample regime. We hypothesize that in this regime, the variance of the estimators makes up the bulk of the error, thus regularization techniques and information in the non-confounding latents could be effectively used to reduce the error as shown in the superior performance of LASSO and ridge regression methods. While, with enough samples, the estimated causal coefficients in the linear regression adjusting with the true, or ICA-estimated, latent confounder, have little variance and small bias and thus exhibit better performances with respect to the LASSO or ridge regressions. Additionally, the GD-PCF causal coefficient estimations also improve upon the baseline methods in this high sample size regime.

\subsection{Climate science example} 
\paragraph{Setting}
In this section, we leverage PCF to uncover teleconnections from high-dimensional spatiotemporal data, a vital aspect of comprehending regional weather and climate dynamics. 
While some studies, e.g., \citep{BAMS-2021-KretschmerAAP}, advocate employing physical knowledge to quantify causal pathways from data, Here, 
we use the PCF framework to identify the common drivers (confounders) responsible for precipitation patterns in both the Mediterranean (MED) and Denmark (DK).
\cite{BAMS-2021-KretschmerAAP} initially observed a significant negative correlation ($r=-0.24$) between precipitation patterns in the MED and DEN.
However, when we rigorously control for the NAO, the correlation between MED and DK reduces to $0.01$ (non-significant). This result highlights the paramount role of NAO in shaping precipitation dynamics within these geographical areas, which can be described by the two following linear models \citep{BAMS-2021-KretschmerAAP}: $\text{DK} = -0.58~\text{NAO} + \varepsilon_1$ and $\text{MED} = 0.42~\text{NAO} + \varepsilon_2$.

We leverage high-dimensional proxies (geopotential height) to uncover the common driver of precipitation in DK and MED. Although we do not anticipate a direct causal link between precipitation in DK and MED, NAO biases the estimation of their causal effects on each other. Acknowledging that our experiment offers a streamlined application of the framework developed in this paper, our intention is to demonstrate one potential application of the framework by focusing on a specific, distilled scenario. Also, given the complexity of climate interactions, we do not assume a single confounder. Instead, we select all confounders that exhibit significant confounding bias. We define the statistical significance of regression models for both MED and DK as those with $p_x + p_y \leq 0.05$. This threshold is chosen because significant $p$-values are generally considered those below $0.05$, and since we are checking two $p$-values, we want their sum to be less than $0.05$. We choose this threshold to use our PCF method to discover new potential confounders in this setting. Future work is to determine a more principled threshold other than $0.05$. Furthermore, since we lack a priori knowledge about the causal direction between MED and DK, we control for MED when predicting DK and vice versa.

In summary, this problem fits into the PCF framework by taking the proxies ($U$) as geopotential height, the treatment ($X$) as precipitation in MED, and the outcome ($Y$) as precipitation in DEN. The PCF framework assumes that relationships between these random variables are linear according to the SCM (Eq.~\eqref{eq:SCM}). Finally, using ICA-PCF in this setting assumes that the latent confounders are sampled from a non-Gaussian distribution. 

\paragraph{Data}  
In this experiment, we employed monthly National Centers for Environmental Prediction (NCEP) \href{https://psl.noaa.gov/data/gridded/data.ncep.reanalysis.html}{reanalyses data} spanning the years $1949$ to $2019$. Our approach for extracting the NAO, MED, and DK signals closely follows the methodology in~\citep{BAMS-2021-KretschmerAAP}. Specifically, we collect monthly signals and focus on the average of June to August to obtain yearly summer means. As proxy variables, we utilize geopotential height data at a pressure level of $500$mb in the Atlantic region ($90$N-$90$S, $280$-$360$E) derived from NCEP reanalyses, spanning the same time frame. This choice was substantiated by the fact that NAO calculations involve a Rotated PCA within the region ($20$N-$90$N)\footnote{NAO computation details are \href{https://www.cpc.ncep.noaa.gov/products/precip/CWlink/daily_ao_index/history/method.shtml}{here}.}.

\paragraph{Numerical comparisons} 

For estimating the AER, we use as a baseline the causal effect estimated by adjusting for the first $10$ principal components of the proxy variables ($0.050 \pm 0.215$, see Table \ref{tab:results_common_driver} column PCA adjustment). ICA-PCF successfully identifies a signal related to NAO with a strong correlation of $0.694$. Employing ICA, we discover $4$ independent components, exhibiting correlations with NAO of $0.694, 0.289, 0.051$, and $0.431$. Furthermore, all PCA-PCF, PLS-PCF, and ICA-PCF  exhibit uncertainty bounds encompassing the true causal effect. Note that uncertainty is notably elevated due to the limited sample size. Combined, the $4$ ICA-PCF components explain a larger portion ($75.9\%$) of the North Atlantic Oscillation (NAO) variance compared to both PCA-PCF and PLS-PCF methods, each of which identifies two confounding components that explain $59.8\%$ and $58.3\%$ of the NAO variance, respectively (see Table \ref{tab:results_common_driver}). The causal graph generated by ICA-PCF is in Fig.~\ref{fig:ICA-PCF-NAO}.
Additional results are shown in Appendix~\ref{app:commondriver}.

\begin{table}[ht!]
    \centering
    \caption{Performance of PCF for identifying common precipitation drivers in Denmark and the Mediterranean. AbsCor and AER only use the $1^{\text{st}}$ extracted confounder; MSE NAO is the Mean Squared Error for NAO prediction with all identified confounders. At the same time, CE is the estimated causal effect adjusting for all PCF confounders ($95\%$ uncertainty).}
    \begin{tabular}{lcccc}
        \toprule
            {} & PCA-PCF & PLS-PCF & ICA-PCF & PCA adjustment \\
        \midrule
            AbsCor & $0.747$ & $0.808 $& $0.694$ & -- \\
            AER & $0.224$ & $1.622$ & $0.815$ & -- \\
            MSE NAO & $0.402$ & $0.417$ & $0.241$ & -- \\
            CE & $-0.005 \pm 0.190$ & $0.094$ $\pm$ $0.196$ & $0.037$ $\pm$ $0.207$ & $0.050 \pm 0.215$ \\
        \bottomrule
    \end{tabular}
    \label{tab:results_common_driver}
    \end{table}

\begin{figure}[ht!]
    \centering
    \begin{tikzpicture}[node distance=3cm]
      \node[draw, circle, line width=1.5pt] (A1) at (0, 3) {$\hat{\mathbf{z}}^{(1)}_c$};
      \node[draw, circle, line width=1.5pt] (A2) at (2, 3) {$\hat{\mathbf{z}}^{(2)}_c$};
      \node[draw, circle, line width=1.5pt] (A3) at (4, 3) {$\hat{\mathbf{z}}^{(3)}_c$};
      \node[draw, circle, line width=1.5pt] (A4) at (6, 3) {$\hat{\mathbf{z}}^{(4)}_c$};
      \node[draw, circle, line width=1.5pt] (B1) at (1, 0) {MED};
      \node[draw, circle, line width=1.5pt] (B2) at (5, 0) {DEN};
      
      \draw[->, red!80!black, line width=1.5pt] (A1) -- (B1) node[midway, left] {0.21};
      \draw[->, blue!80!black, line width=1.5pt] (A1) -- (B2) node[near start, left] {-0.37};
      \draw[->, blue!80!black, line width=1.5pt] (A2) -- (B1) node[near start, right] {-0.4};
      \draw[->, red!80!black, line width=1.5pt] (A2) -- (B2) node[midway, left] {0.4};
      \draw[->, blue!80!black, line width=1.5pt] (A3) -- (B1) node[near start, right] {-0.04};
      \draw[->, red!80!black, line width=1.5pt] (A3) -- (B2) node[midway, left] {0.29};
      \draw[->, blue!80!black, line width=1.5pt] (A4) -- (B1) node[near start, right] {-0.19};
      \draw[->, red!80!black, line width=1.5pt] (A4) -- (B2) node[midway, left] {0.31};
    \end{tikzpicture}
    \caption{ICA-PCF detects $4$ latent confounders ($\hat{\mathbf{z}}^{(1)}_c$, $\hat{\mathbf{z}}^{(2)}_c$, $\hat{\mathbf{z}}^{(3)}_c$, $\hat{\mathbf{z}}^{(4)}_c$) from geopotential height. Their respective correlations with NAO are $0.69$, $-0.28$, $0.05$, and $-0.45$. The edges are the estimated causal effect (regression coefficients).}
    \label{fig:ICA-PCF-NAO}
\end{figure}
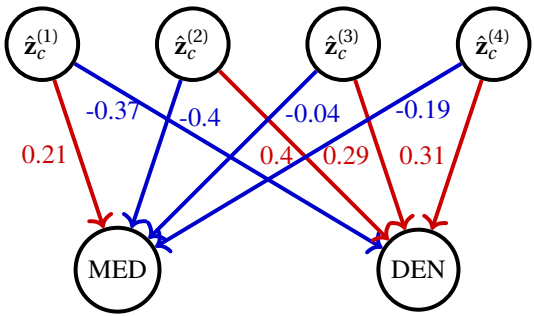

\begin{figure}[t!]
    \centering
    \begin{tikzpicture}[auto,  
    edge/.style ={arrows=-{Latex[length=2mm]}}]
        \node (figa) at (0,0) {\includegraphics[width=0.45\textwidth]{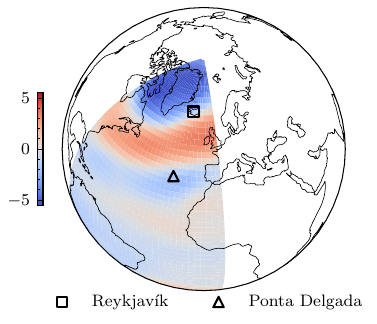}};
        \node (a) at (-2.5,2.5) {(a)};
        \node (figb) at (6.5,0) {\includegraphics[width=0.45\textwidth]{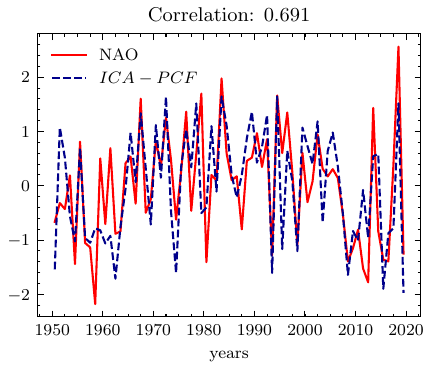}};
        \node (b) at (3,2.5) {(b)};

        \node[circle, draw, fill=lightgray, inner sep=0.4pt, minimum size = 16pt] (dk) at (0.8, 0.6) {\tiny \textbf{DK}};

        \node[circle, draw, fill=lightgray, minimum size = 1pt, inner sep=0.1pt] (med) at (0.9, -0.2) {\tiny \textbf{MED}};

        \node[circle, draw, fill=lightgray, minimum size = 16pt, inner sep=0.1pt] (nao) at (-0.1, 0.2) {\tiny \textbf{NAO}};

            \path[edge, color = black, line width=1pt] (nao) edge (dk);

            \path[edge,
            color = black, line width=1pt] (nao) edge (med);
        
    \end{tikzpicture}
    \caption{(a) Coefficient matrix $\mathbf{A}$ in the ICA-PCF projection, indicating positive and negative associations with the NAO for the Ponta Delgada  (triangle) and 
    Reykjavík meteorological stations (square), respectively. 
    (b) The first confounder extracted by ICA-PCF correlates ($0.694$) with NAO.}
    \label{fig:confounder_weights}
\end{figure}
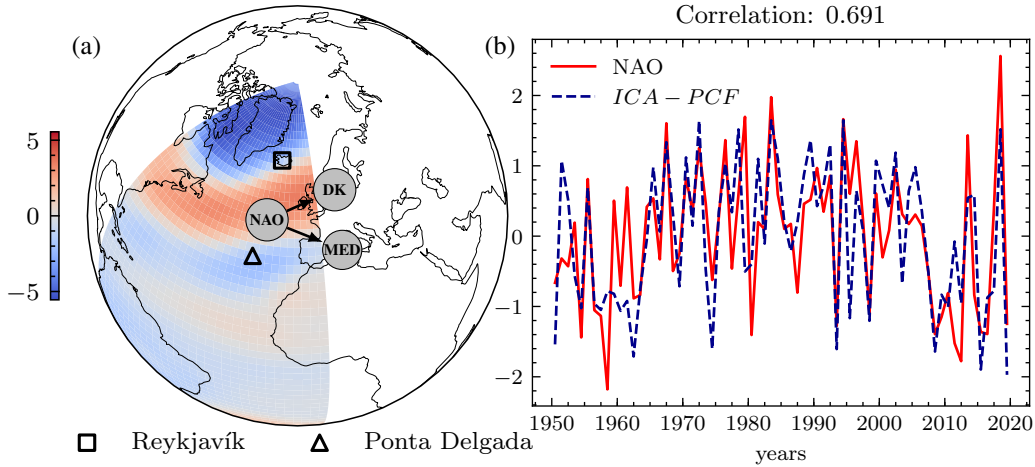

\paragraph{Qualitative comparisons} 
An examination of the ICA projection weights (see Fig.~\ref{fig:confounder_weights}) reveals two distinct patterns, one positive and the other negative. These patterns appear to be related to the geographic locations of meteorological stations (Ponta Delgada and Reykjavík) commonly used for NAO computation. It is worth noting that despite PLS yielding a higher correlation with NAO, it is less effective in mitigating bias in estimating causal effects, and the optimal linear combination of the four components discovered using ICA leads to a lower MSE when trying to predict NAO. However, it is essential to exercise caution when interpreting ICA results, as they exhibited instability, likely due to the limited sample size of $68$ data points, which emphasizes the need for further investigation.

\section{Conclusion}
\paragraph{Summary}
The PCF framework allows us to develop methods for finding low-dimensional confounding variables from high-dimensional proxies and performing accurate adjusted causal effect estimation because it is not restricted by sorted and low-dimensional proxies and the binary treatment case. PCF also admits various dimensionality reduction and selection methods and even end-to-end options like GD-PCF. Our simulations suggest that ICA-PCF and GD-PCF perform best out of all our PCF implementations when the hidden confounders come from a non-Gaussian distribution, and we are in a high sample size regime. Using non-linear regression in the end-to-end GD-PCF approach adapts to the challenging nonlinear setting where dimensionality reduction methods may not identify latent factors as ICA does in the linear case. Additionally, we validated ICA, PLS, and PCA in the PCF framework on an example of high dimensional proxies in climate science and re-discovered known scientific phenomena using ICA-PCF. 

\paragraph{Limitations}
The PCF framework is currently limited to linear mappings and none of our implementations solve the PCF problem when the latent confounders are sampled from a Gaussian distribution. Additionally, the PCF problem needs to be mathematically formalized. Once this is done, the door will be open for theoretical backings for the PCF framework, e.g., consistency theorems. 

\paragraph{Future Work}
In the future, we aim to address these limitations by investigating other implementations of PCF with more sophisticated, nonlinear methods, paying particular attention to methods that can incorporate domain knowledge and (physical) constraints. Additionally, it may be possible to further optimize the parameters in GD-PCF to achieve an optimal end-to-end method for detecting hidden confounders from high-dimensional proxies. Finally, we would like to investigate the theoretical solvability of the difficult case where the hidden confounders come from a Gaussian distribution.

\paragraph{Acknowledgments} 
This work was partly funded by the European Research Council (ERC) under the ERC-CoG-2019 USMILE project (grant agreement 855187) and the Generalitat Valenciana and the Conselleria d’Innovació, Universitats, Ciència i Societat Digital, through the project ``AI4CS: Artificial Intelligence for complex systems: Brain, Earth, Climate, Society'' (CIPROM/2021/56). 

\bibliographystyle{plainnat} 
\bibliography{yourbibfile}

\begin{thebibliography}{40}
\providecommand{\natexlab}[1]{#1}
\providecommand{\url}[1]{\texttt{#1}}
\expandafter\ifx\csname urlstyle\endcsname\relax
  \providecommand{\doi}[1]{doi: #1}\else
  \providecommand{\doi}{doi: \begingroup \urlstyle{rm}\Url}\fi

\bibitem[Arenas-García et~al.(2013)Arenas-García, Petersen, Camps-Valls, and Hansen]{Arenas13}
Jer\'onimo Arenas-García, Kaare~Brandt Petersen, Gustau Camps-Valls, and Lars~Kai Hansen.
\newblock {Kernel multivariate analysis framework for supervised subspace learning: A tutorial on linear and kernel multivariate methods}.
\newblock \emph{IEEE Signal Processing Magazine}, 30\penalty0 (4):\penalty0 16--29, 2013.
\newblock URL \url{http://dx.doi.org/10.1109/MSP.2013.2250591}.

\bibitem[Bradbury et~al.(2018)Bradbury, Frostig, Hawkins, Johnson, Leary, Maclaurin, Necula, Paszke, Vander{P}las, Wanderman-{M}ilne, and Zhang]{jax2018github}
James Bradbury, Roy Frostig, Peter Hawkins, Matthew~James Johnson, Chris Leary, Dougal Maclaurin, George Necula, Adam Paszke, Jake Vander{P}las, Skye Wanderman-{M}ilne, and Qiao Zhang.
\newblock {JAX}: composable transformations of {P}ython+{N}um{P}y programs, 2018.
\newblock URL \url{http://github.com/google/jax}.

\bibitem[Cheng et~al.(2022{\natexlab{a}})Cheng, Li, Liu, Le, Liu, and Yu]{cheng2022sufficient}
Debo Cheng, Jiuyong Li, Lin Liu, Thuc~Duy Le, Jixue Liu, and Kui Yu.
\newblock Sufficient dimension reduction for average causal effect estimation.
\newblock \emph{Data Mining and Knowledge Discovery}, 36\penalty0 (3):\penalty0 1174--1196, 2022{\natexlab{a}}.

\bibitem[Cheng et~al.(2022{\natexlab{b}})Cheng, Liao, Liu, Ma, Xu, and Zheng]{cheng2022learning}
Mingyuan Cheng, Xinru Liao, Quan Liu, Bin Ma, Jian Xu, and Bo~Zheng.
\newblock Learning disentangled representations for counterfactual regression via mutual information minimization.
\newblock In \emph{Proceedings of the 45th International ACM SIGIR Conference on Research and Development in Information Retrieval}, pages 1802--1806, 2022{\natexlab{b}}.

\bibitem[Chu et~al.(2021)Chu, Sun, Dong, Shi, and Huang]{chu2021learning}
Jiebin Chu, Zhoujian Sun, Wei Dong, Jinlong Shi, and Zhengxing Huang.
\newblock On learning disentangled representations for individual treatment effect estimation.
\newblock \emph{Journal of Biomedical Informatics}, 124:\penalty0 103940, 2021.

\bibitem[Comon(1994)]{SP-1994Comon}
Pierre Comon.
\newblock Independent component analysis, a new concept?
\newblock \emph{Signal processing}, 36\penalty0 (3):\penalty0 287--314, 1994.

\bibitem[Cui et~al.(2023)Cui, Pu, Shi, Miao, and Tchetgen~Tchetgen]{JASA-2023CuiPSM}
Yifan Cui, Hongming Pu, Xu~Shi, Wang Miao, and Eric Tchetgen~Tchetgen.
\newblock Semiparametric proximal causal inference.
\newblock \emph{Journal of the American Statistical Association}, pages 1--12, 2023.

\bibitem[Deng et~al.(2012)Deng, Liu, Hu, and Guo]{PR-2012DengLH}
Weihong Deng, Yebin Liu, Jiani Hu, and Jun Guo.
\newblock The small sample size problem of {ICA} : {A} comparative study and analysis.
\newblock \emph{Pattern Recognition}, 45\penalty0 (12):\penalty0 4438--4450, 2012.

\bibitem[Deshpande et~al.(2022)Deshpande, Wang, Sreenivas, Li, and Kuleshov]{deshpande2022deep}
Shachi Deshpande, Kaiwen Wang, Dhruv Sreenivas, Zheng Li, and Volodymyr Kuleshov.
\newblock Deep multi-modal structural equations for causal effect estimation with unstructured proxies.
\newblock \emph{Advances in Neural Information Processing Systems}, 35:\penalty0 10931--10944, 2022.

\bibitem[Díaz et~al.(2023)Díaz, Varando, Johnson, and Camps-Valls]{Diaz_2023}
Emiliano Díaz, Gherardo Varando, J~Emmanuel Johnson, and Gustau Camps-Valls.
\newblock Learning latent functions for causal discovery.
\newblock \emph{Machine Learning: Science and Technology}, 4\penalty0 (3):\penalty0 035004, jul 2023.
\newblock \doi{10.1088/2632-2153/ace151}.
\newblock URL \url{https://dx.doi.org/10.1088/2632-2153/ace151}.

\bibitem[Franklin et~al.(2015)Franklin, Eddings, Glynn, and Schneeweiss]{AJE-2015-FranklinEGS}
Jessica~M Franklin, Wesley Eddings, Robert~J Glynn, and Sebastian Schneeweiss.
\newblock Regularized regression versus the high-dimensional propensity score for confounding adjustment in secondary database analyses.
\newblock \emph{{American Journal of Epidemiology}}, 182\penalty0 (7):\penalty0 651--659, 2015.

\bibitem[Geladi and Kowalski(1986)]{Elsevier-1986GeladiK}
Paul Geladi and Bruce~R Kowalski.
\newblock Partial least-squares regression: {A} tutorial.
\newblock \emph{Analytica chimica acta}, 185:\penalty0 1--17, 1986.

\bibitem[Glass et~al.(2013)Glass, Goodman, Hern\'{a}n, and Samet]{glass2013causal}
Thomas~A. Glass, Steven~N. Goodman, Miguel~A. Hern\'{a}n, and Jonathan~M. Samet.
\newblock Causal inference in public health.
\newblock \emph{Annual Review of Public Health}, 34\penalty0 (1):\penalty0 61--75, 2013.
\newblock \doi{10.1146/annurev-publhealth-031811-124606}.
\newblock URL \url{https://doi.org/10.1146/annurev-publhealth-031811-124606}.
\newblock PMID: 23297653.

\bibitem[Greenfeld and Shalit(2019)]{Greenfeld2019}
Daniel Greenfeld and Uri Shalit.
\newblock Robust learning with the hilbert-schmidt independence criterion.
\newblock In \emph{International Conference on Machine Learning}, 2019.
\newblock URL \url{https://api.semanticscholar.org/CorpusID:203610308}.

\bibitem[Gretton et~al.(2005)Gretton, Bousquet, Smola, and Sch\"{o}lkopf]{HSIC}
Arthur Gretton, Olivier Bousquet, Alex Smola, and Bernhard Sch\"{o}lkopf.
\newblock Measuring statistical dependence with hilbert-schmidt norms.
\newblock In \emph{Proceedings of the 16th International Conference on Algorithmic Learning Theory}, ALT'05, page 63–77, Berlin, Heidelberg, 2005. Springer-Verlag.
\newblock ISBN 354029242X.
\newblock \doi{10.1007/11564089_7}.
\newblock URL \url{https://doi.org/10.1007/11564089_7}.

\bibitem[Hassanpour and Greiner(2019)]{hassanpour2019learning}
Negar Hassanpour and Russell Greiner.
\newblock Learning disentangled representations for counterfactual regression.
\newblock In \emph{International Conference on Learning Representations}, 2019.

\bibitem[Hern\'an and Robins(2023)]{hernan2023causal}
Miguel~A. Hern\'an and James~M. Robins.
\newblock \emph{Causal Inference}.
\newblock Chapman \& Hall/CRC Monographs on Statistics \& Applied Probab. CRC Press, 2023.
\newblock ISBN 9781420076165.
\newblock URL \url{https://books.google.es/books?id=_KnHIAAACAAJ}.

\bibitem[Hotelling(1933)]{JEP-1933Hotelling}
Harold Hotelling.
\newblock Analysis of a complex of statistical variables into principal components.
\newblock \emph{Journal of educational psychology}, 24\penalty0 (6):\penalty0 417, 1933.

\bibitem[Hyv{\"a}rinen and Oja(2000)]{hyvarinen2000independent}
Aapo Hyv{\"a}rinen and Erkki Oja.
\newblock Independent component analysis: algorithms and applications.
\newblock \emph{Neural networks}, 13\penalty0 (4-5):\penalty0 411--430, 2000.

\bibitem[Jutten and Herault(1991)]{jutten1991blind}
Christian Jutten and Jeanny Herault.
\newblock Blind separation of sources, part i: An adaptive algorithm based on neuromimetic architecture.
\newblock \emph{Signal processing}, 24\penalty0 (1):\penalty0 1--10, 1991.

\bibitem[Kretschmer et~al.(2021)Kretschmer, Adams, Arribas, Prudden, Robinson, Saggioro, and Shepherd]{BAMS-2021-KretschmerAAP}
Marlene Kretschmer, Samantha~V Adams, Alberto Arribas, Rachel Prudden, Niall Robinson, Elena Saggioro, and Theodore~G Shepherd.
\newblock Quantifying causal pathways of teleconnections.
\newblock \emph{Bulletin of the American Meteorological Society}, 102\penalty0 (12):\penalty0 E2247--E2263, 2021.

\bibitem[Kuroki and Pearl(2014)]{Biometrika-2014KurokiP}
Manabu Kuroki and Judea Pearl.
\newblock Measurement bias and effect restoration in causal inference.
\newblock \emph{Biometrika}, 101\penalty0 (2):\penalty0 423--437, 2014.

\bibitem[Lee and Lee(1998)]{Springer-1998LeeL}
Te-Won Lee and Te-Won Lee.
\newblock \emph{Independent component analysis}.
\newblock Springer, 1998.

\bibitem[Louizos et~al.(2017)Louizos, Shalit, Mooij, Sontag, Zemel, and Welling]{louizos2017causal}
Christos Louizos, Uri Shalit, Joris~M Mooij, David Sontag, Richard Zemel, and Max Welling.
\newblock Causal effect inference with deep latent-variable models.
\newblock \emph{Advances in neural information processing systems}, 30, 2017.

\bibitem[Luo and Zhu(2020)]{luo2020matching}
Wei Luo and Yeying Zhu.
\newblock Matching using sufficient dimension reduction for causal inference.
\newblock \emph{Journal of Business \& Economic Statistics}, 38\penalty0 (4):\penalty0 888--900, 2020.

\bibitem[Mastouri et~al.(2021)Mastouri, Zhu, Gultchin, Korba, Silva, Kusner, Gretton, and Muandet]{ICML-2021MastouriZGK}
Afsaneh Mastouri, Yuchen Zhu, Limor Gultchin, Anna Korba, Ricardo Silva, Matt Kusner, Arthur Gretton, and Krikamol Muandet.
\newblock Proximal causal learning with kernels: {T}wo-stage estimation and moment restriction.
\newblock In \emph{International conference on machine learning}, pages 7512--7523. PMLR, 2021.

\bibitem[Miao et~al.(2018)Miao, Geng, and Tchetgen~Tchetgen]{Biometrika-2018MiaoGT}
Wang Miao, Zhi Geng, and Eric~J Tchetgen~Tchetgen.
\newblock Identifying causal effects with proxy variables of an unmeasured confounder.
\newblock \emph{Biometrika}, 105\penalty0 (4):\penalty0 987--993, 2018.

\bibitem[Pearl(2010)]{Pearl10}
Judea Pearl.
\newblock On measurement bias in causal inference.
\newblock In \emph{Proceedings of the Twenty-Sixth Conference on Uncertainty in Artificial Intelligence}, UAI'10, page 425–432, Arlington, Virginia, USA, 2010. AUAI Press.
\newblock ISBN 9780974903965.

\bibitem[Peters et~al.(2017)Peters, Janzing, and Sch{\"o}lkopf]{Peters2017}
Jonas Peters, Dominik Janzing, and Bernhard Sch{\"o}lkopf.
\newblock \emph{Elements of Causal Inference: Foundations and Learning Algorithms}.
\newblock Adaptive Computation and Machine Learning. MIT Press, Cambridge, MA, 2017.
\newblock ISBN 978-0-262-03731-0.
\newblock URL \url{https://mitpress.mit.edu/books/elements-causal-inference}.

\bibitem[Runge et~al.(2015)Runge, Petoukhov, Donges, Hlinka, Jajcay, Vejmelka, Hartman, Marwan, Palu{\v{s}}, and Kurths]{Runge2015}
Jakob Runge, Vladimir Petoukhov, Jonathan~F Donges, Jaroslav Hlinka, Nikola Jajcay, Martin Vejmelka, David Hartman, Norbert Marwan, Milan Palu{\v{s}}, and J{\"u}rgen Kurths.
\newblock Identifying causal gateways and mediators in complex spatio-temporal systems.
\newblock \emph{Nature Communications}, 6:\penalty0 8502, 10 2015.
\newblock \doi{10.1038/ncomms9502}.

\bibitem[Runge et~al.(2023)Runge, Gerhardus, Varando, Eyring, and Camps-Valls]{Nature-2023RungeGVE}
Jakob Runge, Andreas Gerhardus, Gherardo Varando, Veronika Eyring, and Gustau Camps-Valls.
\newblock Causal inference for time series.
\newblock \emph{Nature Reviews Earth \& Environment}, 4\penalty0 (7):\penalty0 487--505, 2023.

\bibitem[Schneider et~al.(2013)Schneider, Deser, Fasullo, and Trenberth]{schnider2013Climate}
David~P. Schneider, Clara Deser, John Fasullo, and Kevin~E. Trenberth.
\newblock Climate data guide spurs discovery and understanding.
\newblock \emph{Eos, Transactions American Geophysical Union}, 94\penalty0 (13):\penalty0 121--122, 2013.
\newblock \doi{https://doi.org/10.1002/2013EO130001}.
\newblock URL \url{https://agupubs.onlinelibrary.wiley.com/doi/abs/10.1002/2013EO130001}.

\bibitem[Shortreed and Ertefaie(2017)]{shortreed2017outcome}
Susan~M Shortreed and Ashkan Ertefaie.
\newblock Outcome-adaptive lasso: variable selection for causal inference.
\newblock \emph{Biometrics}, 73\penalty0 (4):\penalty0 1111--1122, 2017.

\bibitem[Snoek et~al.(2019)Snoek, Mileti{\'c}, and Scholte]{snoek2019control}
Lukas Snoek, Steven Mileti{\'c}, and H~Steven Scholte.
\newblock How to control for confounds in decoding analyses of neuroimaging data.
\newblock \emph{Neuroimage}, 184:\penalty0 741--760, 2019.

\bibitem[Sverdrup and Cui(2023)]{OpenReview-2023SverdrupC}
Erik Sverdrup and Yifan Cui.
\newblock Proximal causal learning of conditional average treatment effects.
\newblock 2023.

\bibitem[Tchetgen~Tchetgen et~al.(2020)Tchetgen~Tchetgen, Ying, Cui, Shi, and Miao]{Arxiv-2020TchetgenYCSM}
Eric~J Tchetgen~Tchetgen, Andrew Ying, Yifan Cui, Xu~Shi, and Wang Miao.
\newblock An introduction to proximal causal learning.
\newblock \emph{arXiv preprint arXiv:2009.10982}, 2020.

\bibitem[Wallace and Gutzler(1981)]{Wallace1981TeleconnectionsIT}
John~M. Wallace and David~S. Gutzler.
\newblock Teleconnections in the geopotential height field during the northern hemisphere winter.
\newblock \emph{Monthly Weather Review}, 109:\penalty0 784--812, 1981.
\newblock URL \url{https://api.semanticscholar.org/CorpusID:124794491}.

\bibitem[Wolter and Timlin(2011)]{wolter2011ElNino}
Klaus Wolter and Michael~S. Timlin.
\newblock {El Niño/Southern Oscillation behaviour since 1871 as diagnosed in an extended multivariate ENSO index (MEI.ext)}.
\newblock \emph{International Journal of Climatology}, 31\penalty0 (7):\penalty0 1074--1087, 2011.
\newblock \doi{https://doi.org/10.1002/joc.2336}.
\newblock URL \url{https://rmets.onlinelibrary.wiley.com/doi/abs/10.1002/joc.2336}.

\bibitem[Wu et~al.(2022)Wu, Yuan, Kuang, Li, Wu, Zhu, Zhuang, and Wu]{wu2022learning}
Anpeng Wu, Junkun Yuan, Kun Kuang, Bo~Li, Runze Wu, Qiang Zhu, Yueting Zhuang, and Fei Wu.
\newblock Learning decomposed representations for treatment effect estimation.
\newblock \emph{IEEE Transactions on Knowledge and Data Engineering}, 35\penalty0 (5):\penalty0 4989--5001, 2022.

\bibitem[Wyss et~al.(2022)Wyss, Yanover, El-Hay, Bennett, Platt, Zullo, Sari, Wen, Ye, Yuan, et~al.]{PDS-2022-WyssYCB}
Richard Wyss, Chen Yanover, Tal El-Hay, Dimitri Bennett, Robert~W Platt, Andrew~R Zullo, Grammati Sari, Xuerong Wen, Yizhou Ye, Hongbo Yuan, et~al.
\newblock Machine learning for improving high-dimensional proxy confounder adjustment in healthcare database studies: An overview of the current literature.
\newblock \emph{Pharmacoepidemiology and Drug Safety}, 31\penalty0 (9):\penalty0 932--943, 2022.

\end{thebibliography}

\appendix
\label{sec:appendix}

\section{Dimensionality Reduction}\label{app:dr} 
For the synthetic example, we use the R packages {\tt prcomp} implementation of PCA, {\tt fastICA} package for ICA, and {\tt mdatools} package for PLS.
For the common driver example, we use Python packages {\tt Scikit-learn} implementations of PCA, {\tt PLSSVD} for PLS regression, and {\tt FastICA} for ICA.

PCA was chosen due to its widespread usage, which serves as a sensible baseline. Additionally, PCA is the standard approach for extracting principal modes of variability in climate data~\citep{wolter2011ElNino, schnider2013Climate}. Specifically, for PCA we define the covariance matrix of $\mathbf{U}$ as $\mathbf{C}_{\mathbf{U}}$. PCA finds the orthogonal weights as columns of $\widehat{\mathbf{V}} \in \mathbb{R}^{p \times k}$ via iteratively solving 
\begin{equation*}
    \hat{\mathbf{v}}_{i} = \argmax_{\substack{\mathbf{v} \in \mathbb{R}^{p}, \mathbf{v}_j^T \mathbf{v}_j = 1 \\ \mathbf{v}^T \mathbf{v}_j = 0, \: \forall \: j < i}} \mathbf{v}^T \mathbf{C}_{\mathbf{U}} \mathbf{v} \text{ for }i=1,2\dots, k
\end{equation*}
Then we approximate $\mathbf{Z} \approx \widehat{\mathbf{Z}} =  \mathbf{U} \widehat{\mathbf{V}}$. 

PLS is applied between the proxy and the $(X,Y)$ variables, with the intuition being that we are searching for latent variables highly correlated with both $X$ and $Y$. For our implementation of PLS, we define the matrix $\widetilde{\mathbf{X}} = [\mathbf{X}, \mathbf{Y}]$ and the covariance matrix $\mathbf{C}_{\mathbf{U},\widetilde{\mathbf{X}}}$ between $\mathbf{U}$ and $\widetilde{\mathbf{X}}$. The $i$th column of each of $ \widehat{\mathbf{V}}$ is found by iteratively solving 
\begin{equation*}
    \hat{\mathbf{v}}_{i}, \widetilde{\mathbf{v}}_{i} = \argmax_{\substack{\widetilde{\mathbf{v}}^T \widetilde{\mathbf{v}} = 1 \\ \hat{\mathbf{v}}^T \hat{\mathbf{v}} = 1}} \hat{\mathbf{v}}^T \mathbf{C}_{\mathbf{U},\widetilde{\mathbf{X}}} \widetilde{\mathbf{v}}
\end{equation*}
subject to $\widetilde{\mathbf{v}}^T \widetilde{\mathbf{v}}_j = \hat{\mathbf{v}}^T \hat{\mathbf{v}}_j = 0$ for $j=1,2,\dots,i-1$. Then, similar to PCA, we approximate $\mathbf{Z} \approx \widehat{\mathbf{Z}} =  \mathbf{U} \widehat{\mathbf{V}}$. 

Finally, ICA most closely resembles our scenario because it estimates independent components that are assumed to generate the proxy through a mixing matrix. ICA seeks a coefficient matrix $\mathbf{A} \in \mathbb{R}^{k \times p}$ and a signal matrix $\mathbf{S}\in \mathbb{R}^{n \times k}$ where the columns of $\mathbf{S}$ are independent. Then we approximate $\mathbf{U} \approx  \mathbf{S} \mathbf{A}$. Using this, we approximate $\mathbf{Z} \approx \widehat{\mathbf{Z}} = \mathbf{S}$.




\section{Common driver example}
\label{app:commondriver}

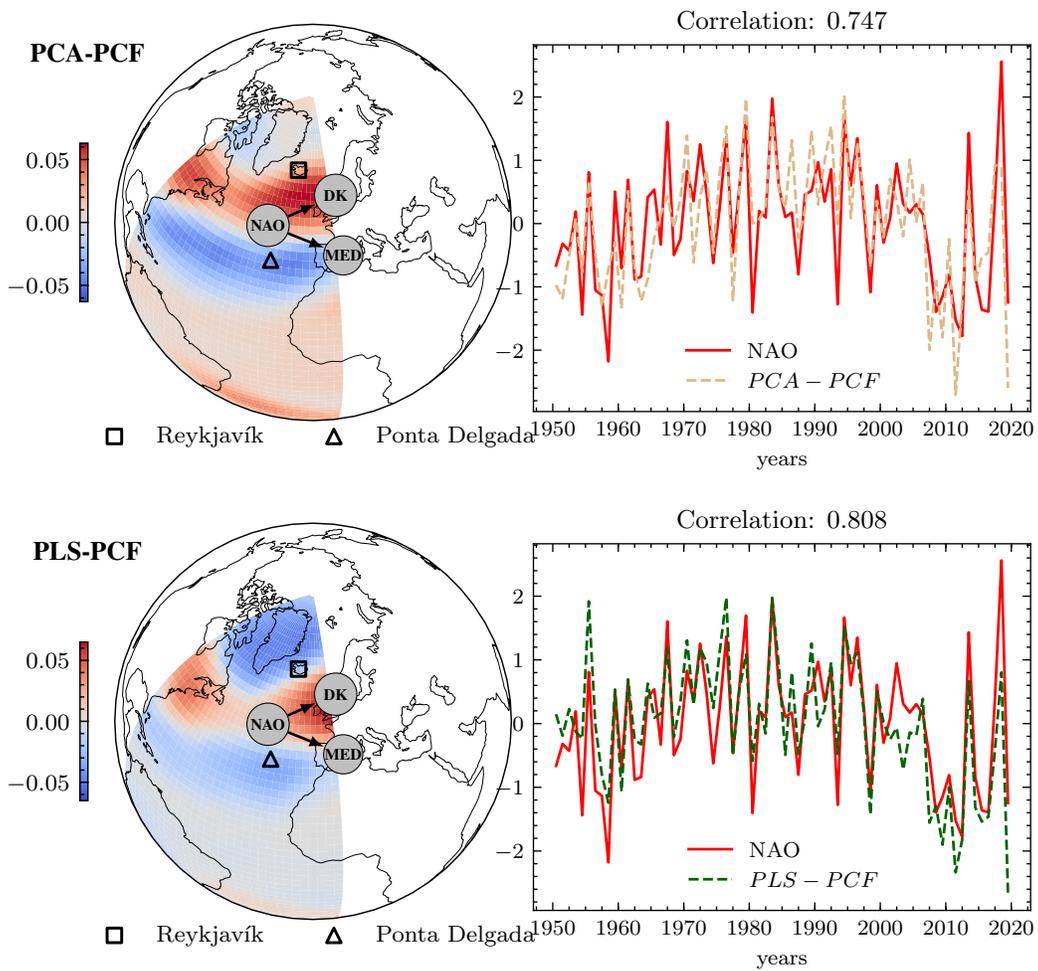
\begin{figure}[ht]
    \centering
    \begin{tikzpicture}[auto,  
    edge/.style ={arrows=-{Latex[length=2mm]}}]
        \node (figa) at (0,0) {\includegraphics[width=0.45\textwidth]{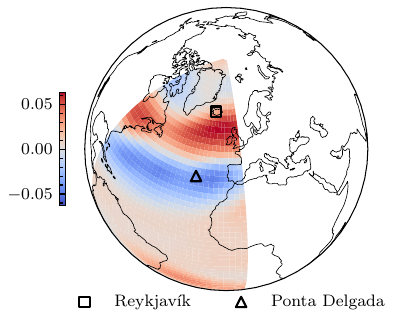}};
        \node (a) at (-2.5,2.5) {\bf PCA-PCF};
        \node (figb) at (6.5,0) {\includegraphics[width=0.45\textwidth]{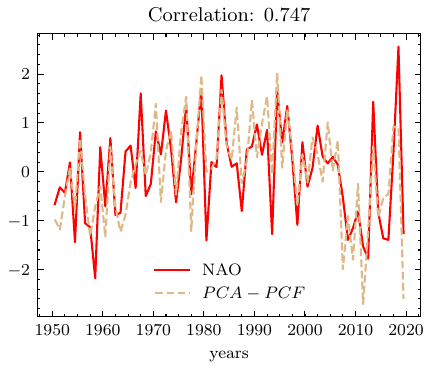}};
        \node (b) at (3,2.5) {};

        \node[circle, draw, fill=lightgray, inner sep=0.4pt, minimum size = 16pt] (dk) at (0.8, 0.6) {\tiny \textbf{DK}};

        \node[circle, draw, fill=lightgray, minimum size = 1pt, inner sep=0.1pt] (med) at (0.9, -0.2) {\tiny \textbf{MED}};

        \node[circle, draw, fill=lightgray, minimum size = 16pt, inner sep=0.1pt] (nao) at (-0.1, 0.2) {\tiny \textbf{NAO}};

            \path[edge, color = black, line width=1pt] (nao) edge (dk);

            \path[edge,
            color = black, line width=1pt] (nao) edge (med);
        
    \end{tikzpicture}

    \begin{tikzpicture}[auto,  
    edge/.style ={arrows=-{Latex[length=2mm]}}]
        \node (figa) at (0,0) {\includegraphics[width=0.45\textwidth]{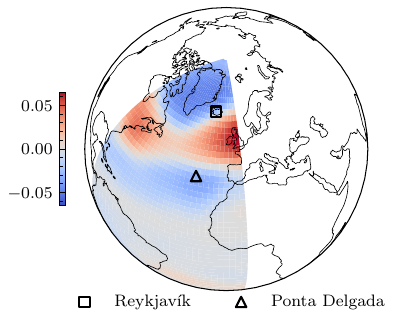}};
        \node (a) at (-2.5,2.5) {\bf PLS-PCF};
        \node (figb) at (6.5,0) {\includegraphics[width=0.45\textwidth]{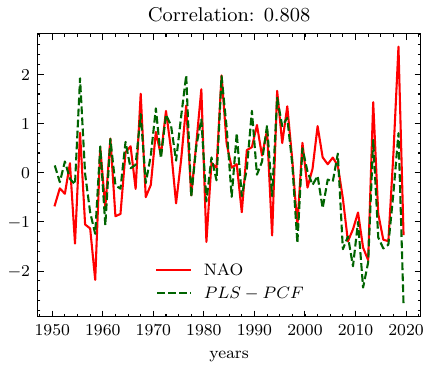}};
        \node (b) at (3,2.5) {};

        \node[circle, draw, fill=lightgray, inner sep=0.4pt, minimum size = 16pt] (dk) at (0.8, 0.6) {\tiny \textbf{DK}};

        \node[circle, draw, fill=lightgray, minimum size = 1pt, inner sep=0.1pt] (med) at (0.9, -0.2) {\tiny \textbf{MED}};

        \node[circle, draw, fill=lightgray, minimum size = 16pt, inner sep=0.1pt] (nao) at (-0.1, 0.2) {\tiny \textbf{NAO}};

            \path[edge, color = black, line width=1pt] (nao) edge (dk);

            \path[edge,
            color = black, line width=1pt] (nao) edge (med);
        
    \end{tikzpicture}
    
    \caption{Coefficient matrices $\mathbf{A}$ (right) and first confounder extracted (left) for the PCA-PCF (top) and PLS-PCF (bottom). We indicate the correlation with NAO.}
    \label{fig:confounder_weights_PCAPLS}
\end{figure}

As shown in Figure \ref{fig:confounder_weights_PCAPLS}, both PCA-PCF and PLS-PCF exhibit similar patterns to ICA-PCF in terms of their weights in the NAO-stations area. It is noteworthy that all three methods fail to accurately recover the NAO signal in the years 2000 to 2005. This may be attributed to the fact that precipitation patterns in both Denmark and the Mediterranean region during this period were not strongly correlated with the NAO. This should be further investigated.

\begin{figure}[ht]
  \centerline{\includegraphics[width=.33\linewidth]{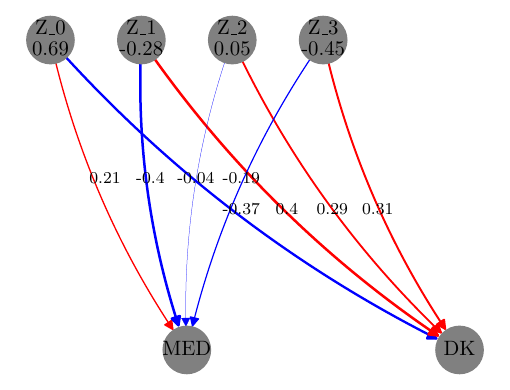}
    \includegraphics[width=.33\linewidth]{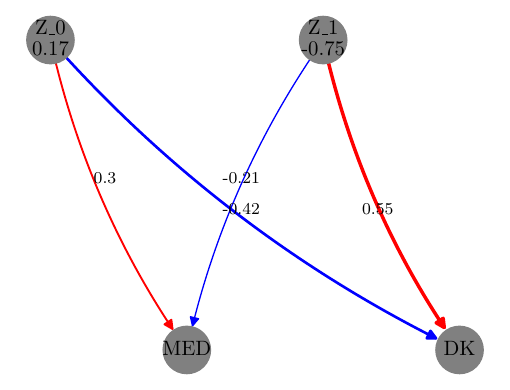}
    \includegraphics[width=.33\linewidth]{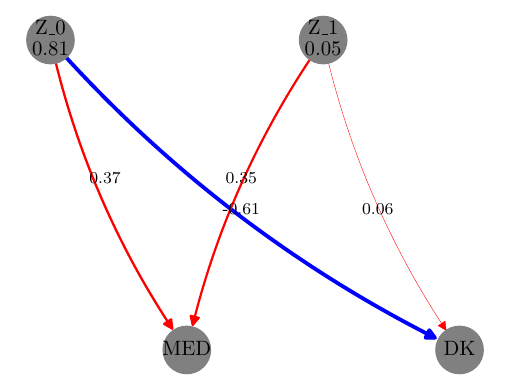}}
    \caption{Causal graph generated by ICA-PCF (left), PCA-PCF (middle), and PLS-PCF (right) with the correlation to NAO indicated beneath each confounder.\label{fig:image1}}
\end{figure}

We provide the linear SCMs induced by each of the three DR algorithms in Fig.~\ref{fig:image1}.  Using ICA in PCF lets us identify direct links from $\{Z_1,Z_2,Z_3\}$ to MED and DK.

\newpage

\end{document}